\documentclass[runningheads]{llncs}
\usepackage{graphicx}
\usepackage{xcolor}
 
\begin{document}
\title{GANCoder: An Automatic Natural Language-to-Programming Language Translation Approach based on GAN}
\titlerunning{GANCoder}
%
\author{Yabing Zhu \and
Yanfeng Zhang \and
Huili Yang \and
Fangjing Wang}
\authorrunning{Y.Zhu et al.}
%
\institute{Northeastern University, China
}
\maketitle        
\begin{abstract}
We propose GANCoder, an automatic programming approach based on Generative Adversarial Networks (GAN), which can generate the same functional and logical programming language codes conditioned on the given natural language utterances. The adversarial training between generator and discriminator helps generator learn distribution of dataset and improve code generation quality. Our experimental results show that GANCoder can achieve comparable accuracy with the state-of-the-art methods and is more stable when programming languages.

\end{abstract}
\setcounter{secnumdepth}{3}
\section{Introduction}
With the development of deep learning and natural language processing (NLP), translation tasks and techniques have been significantly enhanced. The problem of cross-language communication has been well solved. In this digital age, we are no longer a passive receiver of information but also a producer and analyst of data. We need to have data management, query, and analysis skills. Especially, programming is an essential skill in the era of AI. However, it requires strong professional knowledge and practical experience to learn programming languages and write codes to process data efficiently. Although programming languages, such as SQL and Python, are relatively simple, due to education and professional limitations, it is still difficult for many people to learn. How to lower the access threshold of learning programming languages and make coding easier is worth studying. 

In this paper, we explore how to automatically generate programming codes from natural language utterances. The inexperienced users only need to describe what they want to implement in natural language, then the programming codes with the same functionality can be generated via a generator \cite{ref_article1}, so that can simply complete complex tasks, such as database management, programming, and data processing.

Automatic programming is a difficult task in the field of artificial intelligence. It is also a significant symbol of strong artificial intelligence. Many researchers have been studying how to convert natural language utterances into program code for a long time. Before deep learning is applied, pattern matching was the most popular method. But due to the need for a large number of artificial design templates and the diversity and fuzziness of natural language expressions, matching-based methods are not flexible and hard to meet the needs. With the development of machine translation, some researchers try to use statistical machine learning to solve the problem of automatic programming, but due to the difference between the two language models, the results are not satisfactory.

In recent years, GANs have been proposed to deal with the problem of data generation. The game training between GAN's discriminator and generator make the generator learn data distribution better. In this paper, we propose an automatic program generator GANCoder, a GAN-based encoder-decoder framework which realizes the translation between natural language and programming language. In the training phase, we adopt GAN to improve the accuracy of automatic programming generator \cite{ref_article2}. The main contributions of this model are summarized as follows. (1) Introducing GAN into automatic programming tasks, the antagonistic game between GAN's Generator and Discriminator can make Generator learn better distribution characteristics of data; (2) Using Encoder-Decoder framework to achieve the end-to-end conversion between two languages; (3) Using grammatical information of programming language when generating program codes, which provides prior knowledge and template for decoding, and also solves the problem of inconsistency between natural language model and programming language model. Our results show that GANCoder can achieve comparable accuracy with the state-of-the-art methods and is more stable when working on different programming languages.

\section{Related Work}
\subsection{Semantic Parsing and Code Generation}
Semantic parsing is the task of converting a natural language utterance to a logical form: a machine-understandable representation of its meaning,  such as first-order logical representation, lambda calculus, semantic graph, and etc. Semantic parsing can thus be understood as extracting the precise meaning of an utterance  \cite{ref_article3}. Applications of semantic parsing include machine translation, question answering and code generation. We focus on code generation in this paper.

The early semantic analysis systems for code generation are rule-based and limited to specific areas, such as the SAVVY system \cite{ref_article4}. It relies on pattern matching to extract words and sentences from natural language utterances according to pre-defined semantic rules. The LUNAR system \cite{ref_article22} works based on grammatical features, which converts the natural language into a grammar tree with a self-defined parser, and then transforms the grammar tree into an SQL expression. Feature extraction by handcraft not only relies on a large amount of manual work but also impacts the performance of such semantic analysis systems which is relatively fragile. Because the pre-specified rules and semantic templates cannot match the characteristics of natural language expression with ambiguity and expression diversity, the early system functionalities are relatively simple. Only simple semantic analysis tasks are supported. Later, researchers have proposed WASP \cite{ref_article17} and KRISP \cite{ref_article18}, combined with the grammar information of the logical forms, using statistical machine learning and SVM (Support Vector Machine) to convert natural utterances' grammar tree to the grammar tree of the logical forms. Chris Quirk et al. propose a translation approach from natural language utterances to the If-this-then-that program using the KRISP algorithm \cite{ref_article5}.
%
%
%

\begin{figure}
	\begin{center}
		\includegraphics[]{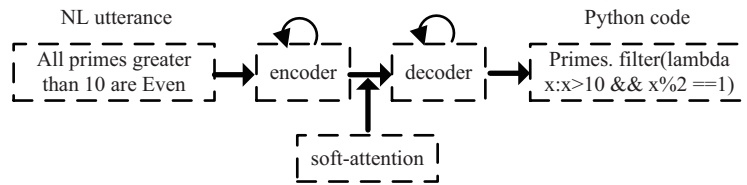}
		\caption{ Encoder-Decoder model for code generation} \label{fig1}
	\end{center}
\end{figure}

Encoder-Decoder frameworks based on RNNs (Recurrent Neural Networks) have been introduced into the code generation tasks. These frameworks have shown state-of-the-art performance in some fields, such as machine translation, syntax parsing, and image caption generation. The use of neural networks can reduce the need for custom lexical, templates, and manual features, and also do not need to produce intermediate representations. Li Dong et al. use the Encoder-Decoder model to study the code generation task and propose a general neural network semantic parser \cite{ref_article3}. Xi Victoria Lin et al. propose an encoder-decoder-based model that converts natural language utterances into Linux shell scripts \cite{ref_article6}. As shown in fig.\ref{fig1}, as an end-to-end learning framework, encoder encodes natural language utterances into intermediate semantic vectors, and decoder decodes intermediate vectors into logical forms. Generally speaking, encoder and decoder can be any neural networks, but LSTM and RNN are mostly used. 

Although programming languages are sequential strings in form, they have a hierarchical structure. A number of works utilize the hierarchical structure property of programs to generate codes .For example, the selective clause and the where clause in SQL belong to different logical levels. Based on this observation, researchers propose tree-based LSTM and tree-based CNN \cite{ref_article7,ref_article8}. Besides, EgoCoder, a hierarchical neural network based on Python program's AST (Abstract Syntax Tree), achieves code auto-completion and code synthesis \cite{ref_article9}. Yin and Neubig propose an Encoder-Decoder model that uses syntax information as the prior knowledge to help decoder reduce search space \cite{ref_article10}. 
%
\vspace{-0.6cm}
\vspace{-1.6cm}
\subsection{Generative Adversarial Network (GAN)}
GAN \cite{ref_article2}, proposed by Ian Goodfellow in 2014, is a method of unsupervised learning. GAN consists of a generator network and a discriminator network. The generator produces what we want, and the discriminator judges whether the output of the generator is fake or subject to the real distribution. Generator and discriminator improve themselves and adjust parameters via their adversarial training.
Since GAN was proposed, it has attracted a lot of attention, and more GANs have been used in image generation, speech synthesis, etc. and achieved much success. CDGAN \cite{ref_article19}, WGAN \cite{ref_article20}, VAEGAN \cite{ref_article21} are typical models of GANs. Since the object processed in NLP is discrete characters, the gradient of discriminator cannot be passed to the generator, so the application of GAN in NLP is not very successful. Lantao Yu et al. proposed SeqGAN \cite{ref_article12} to optimize the GAN network by using the strategy gradient in reinforcement learning to improve the quality of text generation. This is also a successful attempt of GAN in NLP tasks .
\vspace{-0.7cm}
\begin{figure}
	\begin{center}
		\includegraphics[]{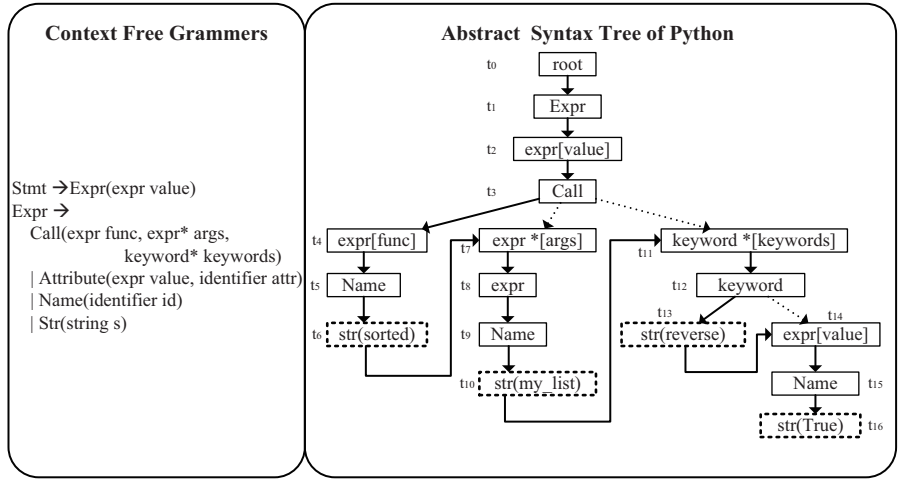}
		\caption{Python context free grammar (left) and abstract syntax tree structure (right)} \label{fig2}
	\end{center}
	
\end{figure}
\vspace{-1.3cm}
\section{Model}

\subsection{GAN-based semantic parsing}
\begin{figure}
	\begin{center}
		\includegraphics[]{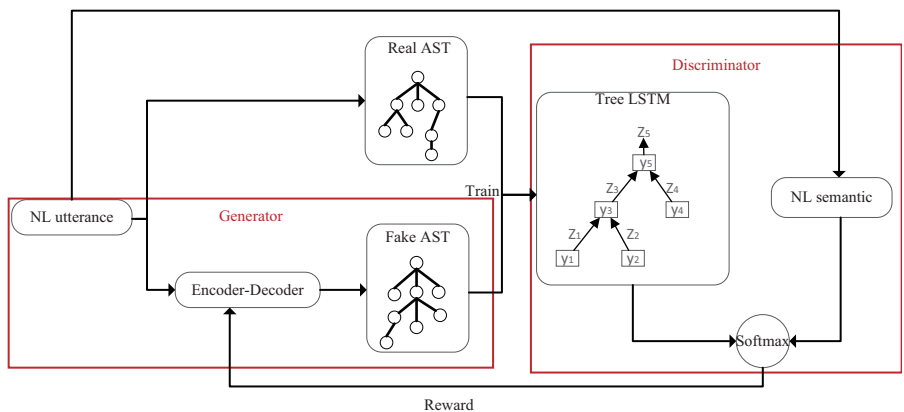}
		\caption{GAN-based automatic program generator} \label{fig3}
	\end{center}
\end{figure}

We aim to design an automatic programming model which can generate a program code sequence  $Y = \{y_1, y_2, \cdots, y_m\} $ based on  a natural language utterance $X = \{x_1, x_2, \cdots, x_n\} $. We introduce GAN into the Encoder-Decoder framework, and propose a new model GANCoder, as shown in fig.\ref{fig3} . In GANCoder, the generator $G_{\theta}$ uses an encoder-decoder framework, which converts the natural language utterances into program codes, where $\theta$ represents parameters. The encoder encodes the natural language utterances as intermediate semantic vectors, and the decoder decodes the semantic vectors into ASTs with the guidance of the programming language grammar information. At last, we parse the ASTs into program codes. The discriminator is responsible for judging whether the ASTs generated by the generator are consistent with the natural language utterance semantics. We use GAN to improve the generative model $G_{\theta}$, the optimization equation of the GAN network is as follows \cite{ref_article13}:
\vspace{-0.1cm}
\begin{equation}
min_{G_\theta}max_{D_\emptyset}\mathcal{L}(\theta, \emptyset) = \mathbf{E}_{X \sim p_x} \log D_\emptyset(X)+ \mathbf{E}_{Y \sim G_\theta} \log(1- D_\emptyset(Y))
\end{equation}
\begin{equation}
G_{\theta} = p(Y|X) = \prod_{t=1}^{|Y|}p(y_t|Y_{<t}, X, Grammar_{CFG})
\end{equation}
where $Y_{<t} =y_1, y_2, \cdots, y_{t-1} $ represents the sequence of the first $t-1$ characters of the program fragment, and $ Grammar_{CFG}$ represents the CFG (Context-Free Grammars) of the programming language, which provides guidance in the code generation process. $X \sim p_x$ indicates that the data sample $X$ is subject to the distribution of real data, and $Y \sim G_\theta$ means that the generator generates the data sample $Y$. Generator and discriminator play two-palyer minimax game. Discriminator can differentiates between the tow distributions.
\vspace{-0.8cm}
\vspace{-1.0cm}
In practice, equation 1 may not provide sufficent gradient for generator when  generating discrete data in NLP. Inspired by the optimization strategy of strategy gradient proposed by SeqGAN, combined with the characteristics of automatic programming tasks, we optimize GANCoder as follows:
\begin{equation}
\mathcal{J(\theta)} =\mathbf{E}_{Y \sim G_\theta} \log (G_\theta(y_1|s_0)\prod_{t=2}^T G_\theta(y_t|Y_{1:t-1}))\mathcal{R}(Y_{1:t})
\end{equation}
\begin{equation}
\mathcal{R}(Y_{1:t}) = D_\emptyset(Y_{1:T})
\end{equation}
where $\mathcal{R}(Y_{1:t})$ represents the generator reward function, which quantifies the quality of the generated program fragments. In the other words, it is the probability of semantic between the natural language utterances and the generated program fragments.

\subsection{CFG-based GAN generator}
\begin{figure}
	\begin{center}
		\includegraphics[width=8cm, height=4cm]{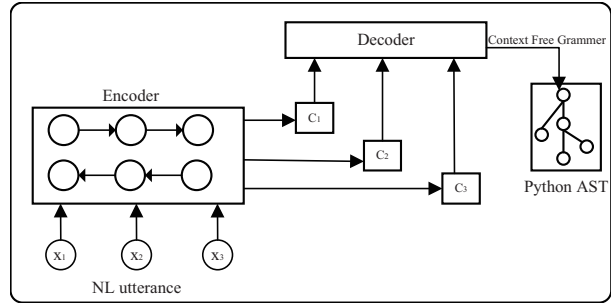}
		\caption{Encoder-Decoder-based generator in GANCoder} \label{fig4}
	\end{center}
\end{figure}

The main task of the GAN generator is to encode the semantics of natural language utterances and then to decode the semantics into AST based on CFG of the programming language. The conversion from one language to another uses the Encoder-Decoder model to reduce the interaction between different languages. The two ends are independently responsible for the processing of their data, simplifying the complexity of the problem. This end-to-end learning framework is more general, and both ends can select their own deep learning models according to the characteristics of the data. Fig.\ref{fig4} shows the framework diagram of the Generator.
\vspace{-0.6cm}
\vspace{-0.8cm}

The encoder is responsible for encoding the semantics of the natural language utterances, as shown in Fig.4. We use a bidirectional LSTM to encode the text sequence of natural language description.$\overleftarrow{h_t}$,and $ \overrightarrow{h_t}$ respectively represent the hidden state of the t-th unit of the natural language description sequence from left to right and from right to left,let $ h_t = [\overleftarrow{h_t}:\overrightarrow{h_t}]$.
be the intermediate hidden vector of the character. The last character's intermediate vector is the semantic presentation of the whole natural utterance.

The decoder decodes the intermediate semantic vector generated by Encoder. Inspired by the model proposed in \cite{ref_article10}, we  first decode the intermediate semantic vector into an abstract syntax tree based on the CFG of the programming language. According to the characteristics of CFG and AST, we define two kinds of actions, corresponding to the generation of non-leaf nodes and leaf nodes in the abstract syntax tree in Fig.\ref{fig2} (right). Logically, we use LSTM to recursively build AST top-down and left-right, as shown in Fig.\ref{fig2} (right). We convert the task  into predicting the grammatic action sequence. Based on the CFG, not only can the template be generated for the decoding process, but also the prediction range can be constrained, so that the search space can be reduced to improve the calculation efficiency.

Unlike encoder, decoder uses a normal LSTM to maintain state variables to form an AST,
\begin{equation}
s_t = f_{LSTM}([a_{t-1}:c_t:p_t], s_{t-1})
\end{equation}
\begin{equation}
p_{action} = Softmax(s_t)
\end{equation}
where "$[:]$" represents the concatenation operation between multiple vectors, $s_t$ represents the state vector at time $t$ in the decoder, The probability of different grammatical actions can be calculated using the $Softmax(s_t)$  function. $a_{t-1}$ represents the vector of the previous action, and $ c_t$ represents the state based on the input $h_X = \{h_{x_1}, h_{x_2}, \cdots, h_{x_{|X|}}\}$. We use the soft-attention mechanism to calculate the attention, as shown in Fig.\ref{fig4}. Then the decoder predicts the probability of each action by the state at time $t$. When a character is generated during a predicted action, we use PointNet to copy the character from the natural language description to AST \cite{ref_article11}. In the process of constructing an AST, the Beam Search algorithm is used to aviod over-fitting. As shown in Fig.\ref{fig2} (right), $t_i$ represents the step of decoding. The order in which nodes are generated is also clearly marked in Fig.\ref{fig2} (right).

\subsection{Tree-based Semantic GAN discriminator}
The Generator can generate an AST of the program fragments, and the discriminator quantifies the similarity of the semantic relationship between the generated ASTs and the natural language utterances. How to quantify the semantic similarity between two different languages is very difficult. In the discriminator, the encoding of natural language utterances still uses the same encoder method in the generator, which uses a bidirectional LSTM to encode the entire sequence into intermediate semantic vectors. When encoding the semantics of a program, there are two different ways. The first is to treat the program code sequence as a string, and still use the same method as the generator processint it in a bidirectional LSTM. The processing is simple, but the logic and syntax information of the program cannot be captured. The second method is to use the structure of the AST  generated by the generator to encode the semantics of the program. However, the semantics of the encoding program is somewhat different from the generation of the AST. In the generator, the structure of the AST is generated recursively top-down and left-right, but in the discriminator, the entire AST is encoded bottom-up from leaf node to root nodes of AST, and the final vector is used as the semantic vector of the program fragment. In this way, the syntax and logic of the program fragment can be learned in a bottom-up manner \cite{ref_article13,ref_article14}.

Let $h_r$  be the final encoding vector for the entire abstract syntax tree. Then the $h_r$ and the semantic vector $h_{NL}$ of the natural language description are classified into two categories:
\vspace{-0.2cm}
\begin{equation}
out=h_r W^{dis} h_{NL}+b^{dis}
\end{equation}
\vspace{-0.5cm}
\begin{equation}
P_{sim}=softmax(out) = \frac{e^{sim}}{\sum_{e \in out} e^i}
\end{equation}
where $P_{sim} \in [0,1]$ represents the probability that the AST is consistent with the semantics of the natural language description. Since the AST is generated based on the CFG of the programming language, the AST is grammatically standardized. The semantics of the generated program fragments need to be consistent with the natural language semantics.


\section{Experiments}
The experiment and evaluation are carried out in a single machine environment. The specific hardware configuration is: processor Intel-i7-8700, memory 32GB, NVIDIA GTX1080 graphics card, memory 8GB. The software environment is: Ubuntu16.04, Python2.7, pytorch3, cuda9. Natural language, Python program characters, and context-free grammar characters embedding are initialized by the xavier\_uniform method \cite{ref_article23}. The optimization function of the model is Adam.

\subsection{Datasets}
\begin{enumerate}
	\item Django is a web framework for Python, where each line of code is manually labeled with the corresponding natural language description text. In this paper, there are 16,000 training data sets and 1805 verification data sets.
	
	\item Atis is the flight booking system dataset, where the natural language utterances are the user booking inquiries, and the codes are expressed in the form of $\lambda$ calculus. There are 4434 training data sets and 140 test data sets.
	
	\item Jobs is a job query dataset, where the user queries are natural language utterance, and the program codes are expressed in the form of Prolog. There are 500 training data sets and 140 test data sets.
\end{enumerate}

\subsection{Experimental results and analysis}

If the sequence of the generated program is the same as the program sequence of the training data, it means that the generated data is correct, and the correctness of the test set indicates the generation effect of the model. As can be seen from Table \ref{tab1}, regarding the Django and Jobs datasets, the pre-trained GANCoder model improves 2.4\% and 0.72\% over the normal generator, respectively, and improves 2.6\% and 2.93\% over that without pre-training. This demonstrates that when training GANCoder, the pre-training has dramatically improved the model. On the ATIS training set, the normal generator is the best. In this model, GAN crashes, which is related to the training data and the grammar rule details of different logical forms. The natural language description sequence of the Jobs data set is relatively simple, so the accuracy of the model is high. The Python language of the Django dataset has relatively good syntax information, but the logic of the program is also more difficult. The game training of GAN also has a good effect.
 The ATIS dataset is logically difficult, but the grammar information is simple, which cannot provide more details when generating ASTs.

Table \ref{tab2} compares the state-of-the-art code generation models with our GANCoder model presented in this paper. Compared to the traditional Encoder-Decoder models, such as SEQ2SEQ and SEQ2TREE, GANCoder increases the accuracy by 24.6\% and 30.3\% on the Django dataset. These two models work better on the Jobs dataset, but the results on the other two datasets are not satisfactory. The ASN model achieves the best performance on the ATIS dataset but cannot obtain results on the other two datasets. The LPN+COPY model and the SNM+COPY model show good results on the Django dataset, but they also do not have results on the other two datasets. Although the GANCoder proposed in this paper is not always the best compared with the other models, GANCoder can achieve relatively stable and satisfactory results on various datasets and is a promising code generation method worth improving.
\vspace{-0.3cm}
\begin{table}
	\caption{Model accuracy based of three training methods}\label{tab1}
	\setlength{\tabcolsep}{2mm}{
		\begin{center}
			\begin{tabular}{lccc}
				\hline
				Dataset & Generator\_normal & GAN\_without\_pretraining & GAN\_with\_pretraining  \\
				\hline
				Django  & 67.3                    & 67.1                   & 69.7                      \\
				Jobs    & 85.71                   & 83.5                   & 86.43                    \\
				ATIS    & 82.6                    & 81.5                   & 79.23                   \\
				\hline
			\end{tabular}
		\end{center}
	}
\end{table}
\vspace{-1.6cm}
\begin{table}
	\caption{Accuracy comparison of different models}\label{tab2}
	\setlength{\tabcolsep}{7mm} {
		\begin{center}
			\begin{tabular}{lccc}
				\hline
				Models     & ATIS      & Django      & Jobs     \\
				\hline
				SEQ2SEQ\cite{ref_article3}    & 84.2      & 45.1        & 87.1     \\
				SEQ2TREE\cite{ref_article3}   & 84.6      & 39.4        & 90.0    \\
				ASN\cite{ref_article15}        & 85.3      & -           & -    \\
				LPN+COPY\cite{ref_article16}   & -         & 62.3        & -    \\
				SNM+COPY\cite{ref_article10}        & -         & 72.1        & -    \\
				GANCoder(Our model)        & 81.5      & 69.7        & 86.43    \\
				\hline
			\end{tabular}
		\end{center}				
	}
\end{table}

\section{Conclusion}
This paper proposes a semantic programming-based automatic programming method GANCoder. Through the game confrontation training of GAN generator and discriminator, it can effectively learn the distribution characteristics of data and improve the quality of code generation. The experimental results show that the proposed GANCoder can achieve comparable accuracy with the state-of-the art code generation model, and the stability is better. The method proposed in this paper can only realize the conversion between single-line natural language description and single-line code. Future work will study how to convert long natural language description text and multi-line code.
\vspace{-0.3cm}
\subsection*{Acknowledgements}
This work was partially supported by National Key R\&D Program of China (2018YFB1003404), National Natural Science Foundation of China (61672141), and Fundamental Research Funds for the Central Universities (N181605017).

\end{document}